\documentclass[conference,letterpaper]{IEEEtran}
\IEEEoverridecommandlockouts
\usepackage{cite}
\usepackage{amsmath,amssymb,amsfonts}
\usepackage{algorithmic}
\usepackage{graphicx}
\usepackage{textcomp}
\usepackage{xcolor}
\usepackage{booktabs}
\usepackage{url}
\usepackage[none]{hyphenat}
\IEEEoverridecommandlockouts
\usepackage{hyperref}
\makeatletter 
\newcommand{\linebreakand}{%
  \end{@IEEEauthorhalign}
  \hfill\mbox{}\par
  \mbox{}\hfill\begin{@IEEEauthorhalign}
}
\makeatother 

\def\BibTeX{{\rm B\kern-.05em{\sc i\kern-.025em b}\kern-.08em
    T\kern-.1667em\lower.7ex\hbox{E}\kern-.125emX}}
\begin{document}

\title{An XR-Based System for Enhanced Spatial Awareness in Mobile Robot Teleoperation\\
}

\author{
\IEEEauthorblockN{Subhransu Sourav Priyadarshan}
\IEEEauthorblockA{\textit{DIBRIS \& DLCM} \\
\textit{Università degli Studi di Genova}\\
Genova, Italy \\
orcid.org/0009-0004-0471-4745}
\and
\IEEEauthorblockN{Valentina Pericu}
\IEEEauthorblockA{\textit{Robotics Lab}\\
\textit{Leonardo S.p.A}\\
Genova, Italy\\
orcid.org/0009-0000-4661-8293}
\and
\IEEEauthorblockN{Federico Rollo}
\IEEEauthorblockA{\textit{Robotics Lab}\\
\textit{Leonardo S.p.A}\\
Genova, Italy\\
orcid.org/0000-0001-5833-5506}
\and
\linebreakand 
\IEEEauthorblockN{Gianni Viardo Vercelli}
\IEEEauthorblockA{\textit{DIBRIS} \\
\textit{Università degli Studi di Genova}\\
Genova, Italy \\
orcid.org/0000-0002-5032-4228}
\and
\IEEEauthorblockN{Navvab Kashiri}
\IEEEauthorblockA{\textit{Robotics Lab}\\
\textit{Leonardo S.p.A}\\
Genova, Italy\\
orcid.org/0000-0002-1219-2447}
}

\IEEEaftertitletext{%
    \vspace{-0.5em}
    \noindent
    \begin{minipage}{\textwidth}
        \footnotesize
        \copyright\ 2026 IEEE. Personal use of this material
        is permitted. Permission from IEEE must be obtained
        for all other uses, in any current or future media,
        including reprinting/republishing this material for
        advertising or promotional purposes, creating new
        collective works, for resale or redistribution to
        servers or lists, or reuse of any copyrighted
        component of this work in other works.

        \medskip
        Published in the 2026 IEEE/ASME International
        Conference on Advanced Intelligent Mechatronics (AIM).

        DOI: \href{https://doi.org/10.1109/AIM65483.2026.11658231}
        {10.1109/AIM65483.2026.11658231}
    \end{minipage}
    \vspace{1em}
}

\maketitle

\begin{abstract}
Teleoperation of mobile robots represents a critical capability for operations in hazardous environments, yet maintaining adequate situational awareness remains challenging without effective visualization and control interfaces. This work presents a multi-modal extended reality (XR) system for robot teleoperation and scene exploration that integrates immersive visual and haptic feedback modalities. 
To address motion sickness commonly induced by 360-degree camera feeds in head-mounted displays (HMDs), the system employs a fixed dual-camera configuration: a forward-facing camera for primary navigation and a rear-facing camera for enhanced situational awareness without operator discomfort. For navigation in confined or low-visibility conditions, a 16-channel Velodyne LiDAR generates real-time 2D occupancy grid maps, which are converted into vibrational haptic feedback on handheld controllers to provide intuitive, non-visual obstacle detection. User evaluation with 20 participants demonstrates that this haptic modality significantly reduces the cognitive load associated with obstacle avoidance during teleoperation.
\end{abstract}


\section{Introduction}
Mobile robot teleoperation remains a fundamental approach for scenarios where full autonomy is unreliable or human judgment is essential, such as in disaster response, industrial inspection, and exploration of hazardous environments \cite{b1,b2}. XR teleoperation systems leverage real-time sensor data, including visual and spatial information, to construct immersive operator interfaces. By rendering the robot's sensory stream in real time, these systems support timely and safe operator decision-making, a principle also adopted in recent sensor-driven digital twin frameworks for safety-critical monitoring \cite{b3}. While advancements in sensing and autonomy have expanded robotic capabilities, the efficacy of teleoperation relies heavily on the quality of the human-robot interface (HRI). Specifically, the operator's ability to maintain situational awareness of the robot’s immediate surroundings is critical for navigation performance and safety \cite{b4}.
Traditional teleoperation interfaces predominantly rely on visual feedback provided by onboard cameras, visualized on 2D monitors. Although widely adopted, these systems often suffer from a limited field of view (FoV) and insufficient depth perception, making it difficult for operators to accurately assess distances to obstacles or comprehend the global spatial layout \cite{b18}. 

\begin{figure}[htbp]
  \centering
  \includegraphics[width=\columnwidth]{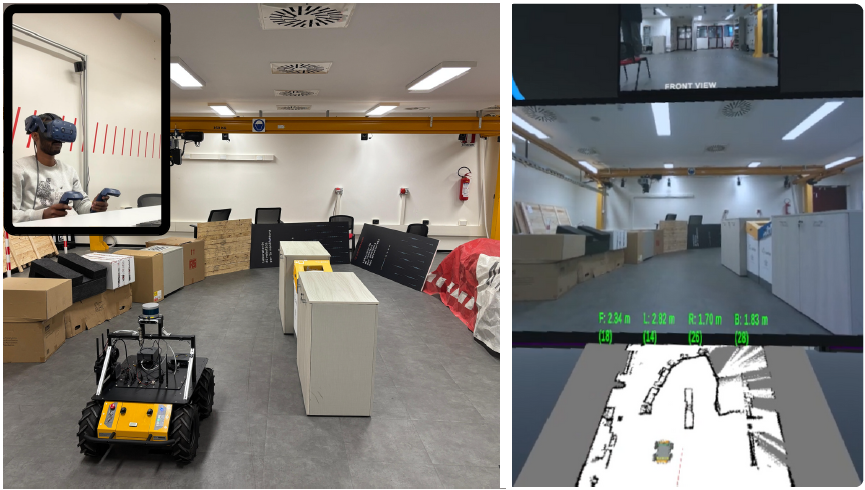}
  \caption{Teleoperative Mobile robot system: a teleoperator with a VR headset, controlling the User interface.}
  \label{fig:front_poster}
\end{figure}

To mitigate these limitations, recent research has transitioned toward immersive Virtual Reality (VR) interfaces. VR-based teleoperation significantly enhances spatial perception and task performance by embedding the operator within a 3D reconstruction of the remote environment, thereby providing a heightened sense of presence compared to conventional screen-based methods \cite{b5,b6}. Studies comparing desktop monitors to VR interfaces consistently demonstrate that the spatially coherent nature of VR leads to superior situational awareness and manipulation performance \cite{b16}.
Existing literature has explored visual enhancement strategies in VR. Approaches utilizing RGB-D cameras and Simultaneous Localization and Mapping (SLAM) techniques allow for the real-time reconstruction of 3D environments \cite{b3,b7}. However, high-fidelity visualizations often introduce significant bandwidth requirements and visual clutter. Furthermore, improper camera handling in VR can lead to severe motion sickness (cyber-sickness). To address these issues, recent dual-mode architectures have been proposed that decouple navigation views from task-specific views, such as manipulation, allowing each viewpoint to be optimized for its respective perceptual requirements \cite{b11}. Specifically, the use of fixed stereoscopic dual-camera feeds has been shown to provide necessary depth perception without the disorientation often caused by 360-degree video streams \cite{b12, b13}.
However, dependence on visual feedback alone presents a significant constraint. In conditions of low visibility, narrow passages, or high visual complexity, operators may not identify nearby obstacles through visual feedback alone. To address this, alternative sensory modalities have been investigated. Haptic feedback has been successfully employed in robotic manipulation to convey contact forces \cite{b9} to the teleoperator and also to facilitate more intuitive navigation by employing HMD devices \cite{b10}. 

While haptic feedback has been widely explored in mobile robot teleoperation, ranging from vibrotactile formation control \cite{b26} to the use of haptic tablets in 2D interfaces \cite{b27}, these conventional systems largely rely on non-immersive interfaces that still demand high visual cognitive load. Prior studies have demonstrated that multi-modal haptic systems significantly improve operator immersion and efficiency compared to visual-only feedback \cite{b15}. More recent advancements have introduced haptics into immersive environments but often rely on cumbersome force-feedback mechanisms applied to the operator's body \cite{b28} or focus exclusively on fine manipulation and grasping \cite{b29,b30}. In contrast, the novelty of our system lies in integrating lightweight, directional haptic feedback within a fully immersive XR environment to achieve sensory substitution. By mapping a real-time LiDAR occupancy grid directly to left and right handheld VR controllers, our architecture provides a continuous, non-visual ``sixth sense" for spatial awareness.Additionally, theoretical frameworks using occupancy grid maps continue to serve as the primary approach for robust environmental representation in these autonomous and semi-autonomous systems \cite{b14}.

To incorporate robust mapping and intuitive user feedback, we propose a multimodal XR-based teleoperation system. 
While recent advancements in teleoperation have relied on SLAM approaches to enhance situational awareness through photorealistic 3D reconstructions, these systems require immense computational hardware (e.g., high-end GPUs) and place the entirety of the spatial processing burden on the operator's visual channel. In contrast, our system proposes a fundamentally different architectural solution: rather than maximizing visual fidelity, we maximize cognitive efficiency. By integrating a dual-camera stream, lightweight 2D occupancy grids with low latency, and LiDAR-driven haptic feedback, our system provides immediate spatial awareness that does not require heavy rendering times and remains robust in low-visibility conditions where RGB-D-based SLAM would fail.

A direct haptic feedback loop maps real-time LiDAR scans to vibrational cues on handheld VR controllers. This enables operators to perceive the proximity of obstacles, reducing visual dependence and mitigating sensory mismatches that contribute to simulator sickness \cite{b13}.
The main contributions of this work are:
\begin{itemize}
\item A human-centred immersive teleoperation interface designed to enhance situational awareness for efficient navigation;
\item  A multimodal perception framework that integrates dual camera streams with real-time 2D occupancy grid mapping, enabling operators to access both local visual feedback and global spatial context simultaneously;
\item A LiDAR-based haptic feedback mechanism that delivers distance-dependent, directionally intuitive vibration cues through handheld VR controllers to enhance obstacle awareness without increasing visual cognitive load.
\end{itemize}
The proposed approach is detailed across the following sections: Section~\ref{sec:sys_arch} introduces the system architecture, followed by the multimodal perception interface in Section~\ref{sec:perception_int}. A user-study evaluating teleoperation usability is described, and the performance is analyzed in Sections~\ref{sec:exp valid} and \ref{sec:performance}.


\section{System Architecture}
\label{sec:sys_arch}
\begin{figure}[htbp]
  \centering
  \includegraphics[width=\columnwidth]{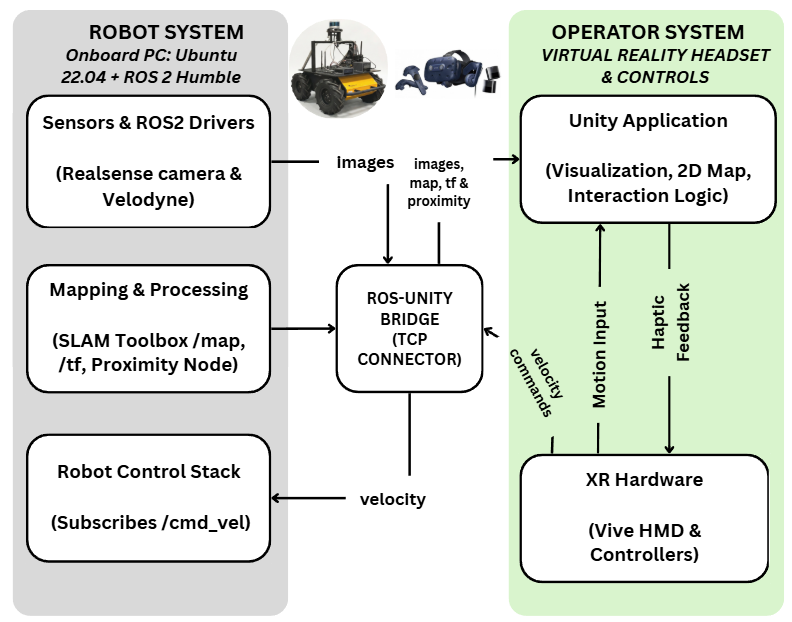}
  \caption{System architecture for a VR-operated mobile robot, showing the ROS-based onboard stack, ROS–Unity TCP bridge, and a Unity XR operator interface with real-time sensor streaming and control feedback.}
  \label{fig:sys_architecture}
\end{figure}

The proposed teleoperation system employs a distributed architecture that isolates robot-side perception and computation from operator-side visualization and control. As illustrated in Fig.~\ref{fig:sys_architecture}, the data flow is structured into three primary modules: (i) the Robot System (sensing and actuation), (ii) the ROS-Unity Bridge (network layer), and (iii) the Operator System (VR visualization and input). 

\subsection{Robot Hardware and Sensing}
The mobile platform is built upon a Clearpath Husky Unmanned Ground Vehicle (UGV) \cite{b20}. Perception is facilitated by a multi-modal sensor suite comprising the components summarized below:
\begin{itemize}
    \item Visual Sensing: The vision system utilizes two Intel RealSense cameras mounted in an opposing configuration: one forward-facing to support primary navigation, and one rear-facing to provide situational awareness and localization support. Each camera streams video at $640 \times 480$ resolution at 30 Hz. Unlike standard monocular setups that limit the operator's view to a single frustum, this dual-view configuration provides comprehensive bi-directional coverage. This allows the operator to reverse safely and maintain rear-guard awareness without the geometric distortions typically associated with $360^{\circ}$ fisheye lenses \cite{b11}.
    \item Volumetric Sensing: For environmental mapping and obstacle detection, the platform integrates a Velodyne VLP-16 3D LiDAR. The sensor provides $360^{\circ}$ horizontal coverage with a vertical Field of View (FoV) sufficient for detecting obstacles at varying elevations (e.g., overhanging objects or low barriers). Operating at a scan frequency of 10 Hz, the LiDAR ensures real-time proximity detection. 
    \item Onboard Computation: All robot-side processing is executed on an onboard PC running Ubuntu 22.04 and ROS 2 Humble \cite{b21}. To ensure system robustness, low-level tasks—including sensor drivers, SLAM processes (SLAM Toolbox), and velocity control interfaces—are executed locally on the robot. This edge-computing approach isolates critical safety functions from potential network instability or latency in the teleoperation link.
\end{itemize}


\subsection{Visual Feedback and Mapping} 
The following describes the visualization and mapping components of the system:
\begin{itemize}
\item Camera-Based Visual Feedback: The representation of the camera feeds and the environmental map within the head-mounted display (HMD) must be carefully designed to support effective navigation. The visual interface subscribes to the dual RGB streams transmitted via ROS\,2 topics. Within the Unity environment \cite{b22}, these feeds are rendered on virtual display surfaces anchored relative to the operator’s viewpoint. Rather than performing computationally expensive stereo depth reconstruction, the system leverages the fixed baseline to provide complementary perspectives. This dual camera setup reduces computing load and eliminates blind spots inherent to single-camera configurations \cite{b12}.
Crucially, the rendering of camera feeds is decoupled from the operator’s head pose. Instead of tracking the user's head movements directly (which can induce sensory conflict), the feeds act as stabilized visual references. This decoupling is a strategic measure to mitigate motion sickness (cybersickness), allowing the operator to scan the environment visually without being constrained to the robot’s instantaneous orientation \cite{b13}.
\item 2D Occupancy Grid Mapping: To establish global spatial context beyond the immediate camera field of view (FoV), the system generates a real-time 2D occupancy grid map derived from LiDAR data. Mapping is handled by the SLAM Toolbox package within the ROS\,2 ecosystem. The SLAM algorithm publishes a standard \texttt{nav\_msgs/Occupancy Grid} message, encoding the probability of free, occupied, and unknown space based on accumulated LiDAR scans and odometry. This map is streamed to Unity and projected as a planar overlay within the virtual environment. The robot’s estimated pose is synchronized with this map in real time, allowing the operator to correlate local visual cues with the global topology. This hybrid visualization addresses the limitations of purely vision-based teleoperation by revealing obstacles that lie outside the active camera frustum \cite{b14,b16}.
\end{itemize} 

\subsection{ROS–Unity Integration}

Data exchange between the robot and the XR interface is managed via a TCP-based ROS–Unity communication bridge. Sensor data (images, LiDAR scans, and odometry) is serialized into standard ROS messages and transmitted to the Unity client, where it is deserialized into native XR assets. In the reverse direction, control inputs from the HTC Vive Pro handheld controllers \cite{b19} are captured by the Unity input system, translated into linear and angular velocity commands, and published to the robot's velocity topic. This bidirectional architecture ensures modularity; perception, mapping, and control subsystems can be developed and validated independently \cite{b9}.


\section{Multimodal Perception Interface}
\label{sec:perception_int}
\begin{figure}[t]
  \centering
  \includegraphics[width=\columnwidth]{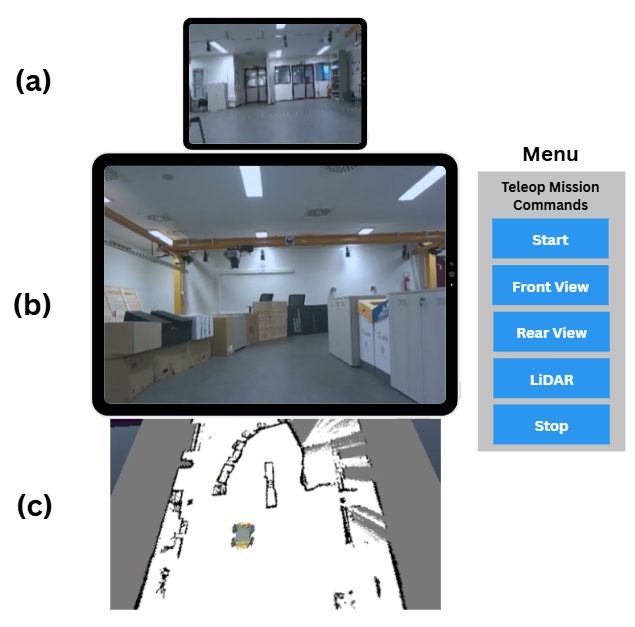}
  \caption{HMD interface (a) Rear Camera View (b) Front Camera View (c) 2D Occupancy Grid Map}
  \label{fig:Unity_GUI}
\end{figure}

The teleoperation interface is designed to mitigate the tunnel vision often associated with camera-only teleoperation. By integrating visual streams with spatial mapping, the system balances immediate maneuverability with global situational awareness through a unified egocentric operating paradigm. 
In Fig.~\ref{fig:Unity_GUI}, the layout shows the front and rear camera views and a 2D occupancy grid map. A menu interface allows operators to switch between modes and enable/disable the camera and/or lidar based on bandwidth requirements.

\subsection{Visual Layout and Interaction} 
The XR environment presents two distinct layers of information. The primary layer consists of live feeds from the dual RealSense cameras, rendered as stabilized virtual panels fixed relative to the robot's forward vector. This fixation ensures that the operator’s visual reference remains consistent with the robot’s heading, reducing the sensory conflict that contributes to simulator sickness \cite{b13}.

\subsection{Global Context Awareness} 
To compensate for the limited FoV of the physical cameras, the system visualizes the generated 2D occupancy grid map as a secondary planar element. We place the map below the operator's main view, positioned similarly to an automotive instrument panel, rather than overlaying it on the video. This prevents visual clutter, allowing the operator to control/navigate the robot using the camera feed while periodically checking the map for hidden obstacles \cite{b14}. In this way, the operator maintains continuous awareness of the robot's global position.

\subsection{LiDAR-Based Haptic Feedback}
While the visual interface addresses spatial orientation, visual load can still become overwhelming in cluttered environments. To offload perceptual demand, the system implements a vibrotactile feedback loop driven by raw LiDAR data. This subsystem functions as a background collision warning mechanism, providing intuitive cues without requiring direct visual attention.
The system computes the minimum Euclidean distance to the nearest obstacle, denoted as $d_{\min}$. 
To provide graded feedback, a linear decay function maps this distance to a vibration intensity $I \in [0,1]$, which is sent to the corresponding HTC Vive Pro handheld controller. 
The vibration intensity is defined as:


\begin{equation} \label{eq:intensity}
    I =
    \begin{cases}
    \dfrac{d_{\text{thresh}} - d_{\min}}{d_{\text{thresh}} - d_{\text{stop}}}, 
    & \text{if } d_{\text{stop}} < d_{\min} < d_{\text{thresh}}, \\[8pt]
    1, 
    & \text{if } d_{\min} \le d_{\text{stop}}, \\[6pt]
    0, 
    & \text{if } d_{\min} \ge d_{\text{thresh}},
    \end{cases}
\end{equation}
where $d_{\min}$ denotes the minimum distance obtained independently for each sector, while $d_{\text{thresh}} = 0.85\,\text{m}$ is the activation threshold at which vibration begins, and $d_{\text{stop}} = 0.50\,\text{m}$ is the safety distance at which the vibration reaches its maximum intensity ($I = 1.0$). The choice of a linear decay function in Equation \ref{eq:intensity} is designed to provide a predictable, proportional escalation in vibration intensity. Because the active warning window is intentionally narrow ($d_{\text{thresh}} = 0.85\,\text{m}$ to $d_{\text{stop}} = 0.50\,\text{m}$), a linear mapping ensures that the operator receives a smooth and steady gradient of urgency as the robot approaches an obstacle. While recent studies in whole-body force-feedback have proposed non-linear mappings, such as sigmoid functions, to prevent the physical discomfort of constant pushing during navigation, our system employs localized vibrotactile feedback. For such vibrotactile cues, linear scaling effectively functions as an intuitive background collision warning without introducing the complexity of non-linear tuning. To resolve this haptic feedback in a way that matches the operator’s egocentric reference frame,
the processed LiDAR point cloud is segmented along the robot’s longitudinal axis, effectively dividing the scan into left and right sectors relative to the robot's base link.  Consequently, the system establishes a direct lateral correspondence: obstacles detected in the left sector modulate the vibration of the left handheld controller, while those in the right sector drive the right controller. Using this direct mapping, the operator can instinctively perceive the direction of a potential collision, such as a wall near the robot's side, using only vibrational feedback, even if the obstacle is outside the camera view.

\subsection{Teleoperation control}

\begin{figure}[htbp]
  \centering
  \includegraphics[width=0.7\columnwidth]{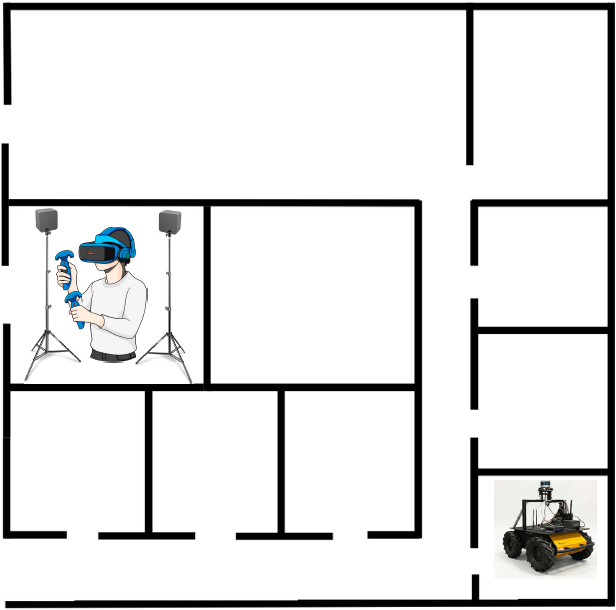}
  \caption{Remote robot operation from another room through an immersive VR setup.}
  \label{fig:controls_layout}
\end{figure}

Control authority is maintained through the HTC Vive Pro handheld controllers using a direct mapping scheme to ensure low-latency response. 
The system is deployed on a Clearpath Husky mobile robot, a differential-drive platform governed by linear ($v$) and angular ($\omega$) velocity commands. As illustrated in Fig.~\ref{fig:controls_layout}, the operator and the mobile robot can be located in different areas. The operator interacts with the robot exclusively through an immersive VR interface, without direct line-of-sight to the physical platform. This separation emphasizes the fully remote nature of the teleoperation setup and eliminates reliance on visual contact with the robot or its environment. By directly translating inputs from the handheld controller into robot velocity commands, the system ensures that control remains intuitive, even though the operator is working remotely.
This arrangement enables safe experimentation and deployment in environments where human presence is impractical or hazardous, while maintaining a strong sense of agency and control for the operator.

User input is captured from the vertical and horizontal axes of the trackpad of the controller held in the operator's dominant hand. 
These analog values are normalized and mapped to the standard ROS \texttt{geometry\_msgs/Twist} message structure. 
To prevent abrupt acceleration that could induce motion sickness or wheel slip, the raw inputs are passed through a software-defined velocity smoothing filter before being published to the \texttt{/cmd\_vel} topic.

The control loop and the haptic feedback loop operate asynchronously. Nevertheless, the operator effectively closes the overall control loop by responding to the vibration cues. 
As the vibration intensity $I$ (see Equation~\ref{eq:intensity}) increases, indicating closer proximity to obstacles, the operator is naturally prompted to reduce the linear velocity $v$ or adjust the angular velocity $\omega$. 
This results in safer navigation behavior without imposing autonomous braking or shared-control overrides by the system \cite{b16}.

\section{Experimental Validation}
\label{sec:exp valid}
To evaluate the efficacy of the proposed multimodal interface, we conducted a comparative user study focusing on navigation performance, safety, and operator workload. The study aimed to quantify the benefits of integrating occupancy grid mapping and haptic feedback into the teleoperation loop.

\subsection{Experimental Setup}

The experiments were conducted in a static indoor environment with a fixed obstacle configuration along the navigation path (see Fig.~\ref{fig:front_poster}). Obstacles of varying height and material were deliberately placed along the route, including low-profile objects and partially occluded panels, to challenge the robot’s perception under limited field-of-view conditions. The layout required precise maneuvering and continuous path adjustment, emphasizing the robot’s ability to detect obstacles, maintain clearance in confined spaces, and safely traverse cluttered indoor environments. During the experiment, the users were equipped with a VR headset and two VR controllers to interact with the robot
A total of 20 volunteers participated in the study. The demographic survey indicated that 55\% of participants had zero prior experience with robot teleoperation, and 55\% had never used VR headsets before. This novice-heavy distribution highlights the system's intuitive design, as no extensive training was provided.

Participants were required to teleoperate the Clearpath Husky robot from a designated start point to a finish point as quickly as possible while minimizing collisions. To evaluate the impact of each sensory channel, each participant performed the navigation task under three distinct modalities:
\begin{itemize}
    \item Mode 1 - baseline: Visual feedback only (front and rear RGB cameras).
    \item Mode 2 - visual augmentation: Dual cameras with real-time 2D occupancy grid map.
    \item Mode 3 - proposed system: Dual cameras with a 2D occupancy grid map and LiDAR-based haptic feedback.
\end{itemize}

\subsection{Quantitative Analysis}

\begin{table}[b]
    \centering
    \caption{Navigation performance metrics averaged over 20 trials}
    \label{tab:nav_performance}
    \begin{tabular}{lcc}
        \toprule
        \textbf{Modality} & \textbf{Mean Time (s)} & \textbf{Avg. Collisions} \\
        \midrule
        \textbf{Mode 1:} Camera Only & 56.0 & 0.75 \\
        \textbf{Mode 2:} Camera + Map & 44.7 & \textbf{0.20} \\
        \textbf{Mode 3:} Camera + Map + Haptics & \textbf{41.9} & 0.30 \\
        \bottomrule
    \end{tabular}
\end{table}

Performance data collected in our experiments is summarized in Table~\ref{tab:nav_performance}.
Two primary metrics were used for our evaluation:
\begin{itemize}
    \item Navigation Efficiency (Time):
The timing data demonstrates a clear trend of improved efficiency with the addition of sensory modalities. Mode 1 yielded the slowest average completion time (Avg: 55.6\,s). Participants frequently exhibited hesitation at corners, indicating a lack of spatial confidence. Mode 2 reduced the average time (Avg: 46.1\,s), an improvement of approximately 17\%. Mode 3 resulted in the fastest performance (Avg: 42.8\,s), representing a 23\% improvement over the baseline. Notably, the variance in completion times decreased in Mode 3, suggesting that the haptic feedback helped level the playing field, allowing novice users (the 55\% majority) to perform closer to the level of experts.
    \item Safety (Collisions): Analysis of the collision data reveals a substantial safety improvement with the introduction of the assisted modalities. Mode 1 (Camera Only) resulted in the highest collision rate, with an average of 0.75 collisions per run. Several participants incurred multiple impacts (up to 6 for one subject), primarily due to side-swipes against obstacles located in the camera's blind spots.
    In contrast, Mode 2 and Mode 3 significantly mitigated this risk. Mode 2 recorded the lowest collision rate (0.20 collisions). This can be attributed to the cognitive nature of the task: interpreting the occupancy map often induced a stop-and-check navigation strategy, where operators intentionally slowed down or paused to verify clearance before committing to a turn. Similarly, the proposed haptic system successfully reduced collision frequency by 60\% compared to the baseline while delivering the fastest completion times. 
    The haptic feedback provided a continuous safety net that encouraged higher navigation speeds (41.9\,s vs 44.7\,s). As a result, operators leveraged the intuitive nature of the haptic cues to drive more aggressively, trading a fraction of the safety margin for significant gains in efficiency. However, Mode 3 showed a slightly higher collision rate (0.30 collisions) that reflects a shift in control strategy. 
\end{itemize}

\begin{figure}[t]
  \centering
  \includegraphics[width=\columnwidth]{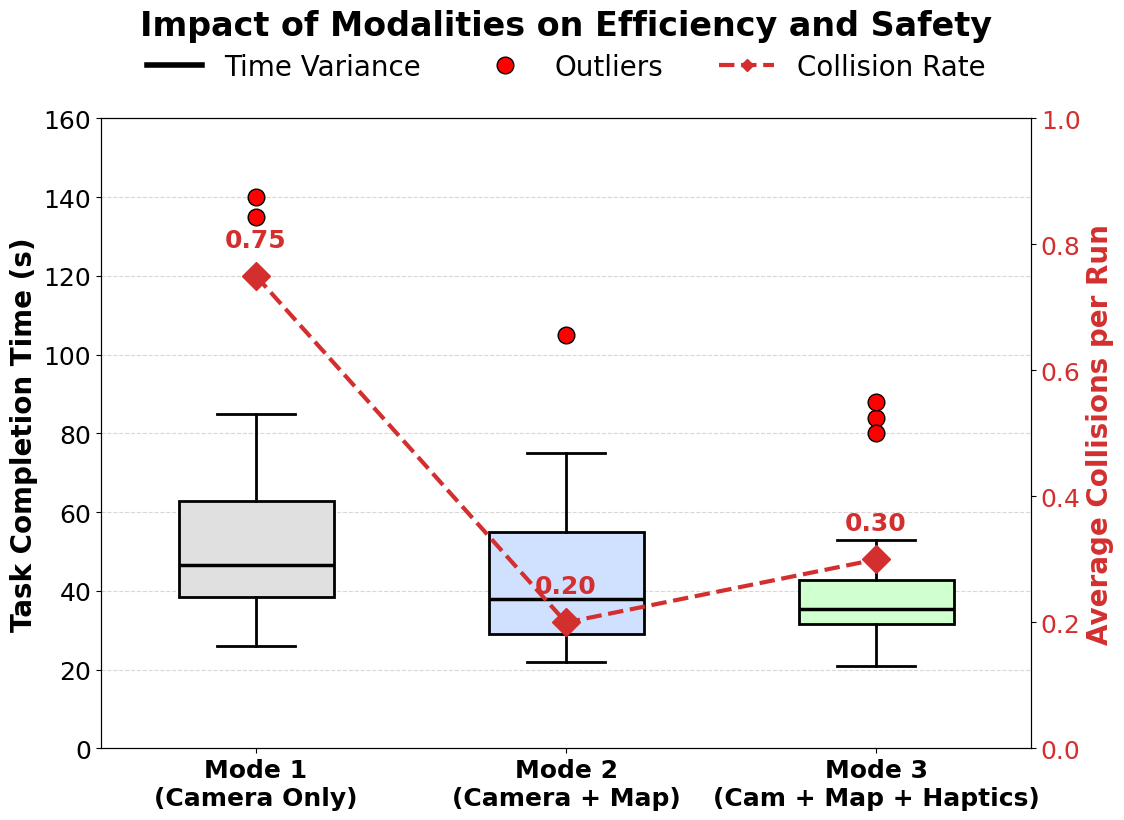}
  \caption{Joint analysis of navigation efficiency and safety. The box plots (left axis) illustrate the distribution of completion times, highlighting the reduction in variance and mean time in Mode 3. The dashed red line (right axis) depicts the average collision rate, demonstrating a sharp safety improvement in the assisted modalities compared to the baseline.}
  \label{fig:graph}
\end{figure}
\subsection{Qualitative Analysis (User Study)}
\begin{figure*}[t]
    \centering
    \includegraphics[width=0.75\textwidth]{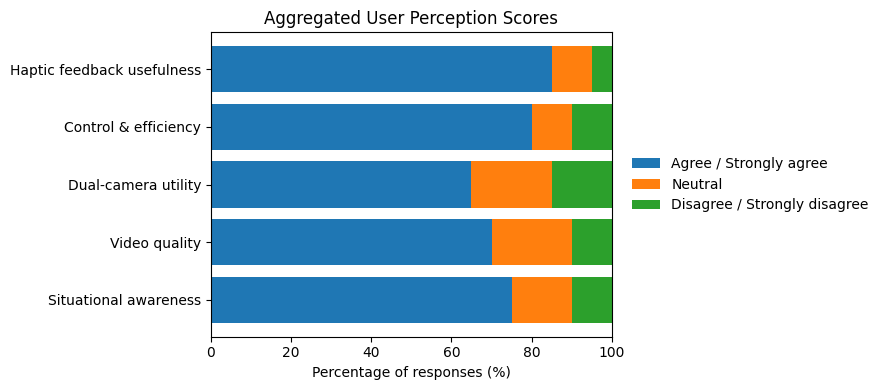}
    \caption{Aggregated user perception scores.
Responses are grouped into agreement, neutral, and disagreement categories on a 7-point Likert scale. Agreement (Agree / Strongly Agree) responses are shown in blue, neutral responses in orange, and disagreement (Disagree / Strongly Disagree) responses in green. The distribution highlights strong user agreement regarding situational awareness, control efficiency, and haptic feedback usefulness, with slightly higher neutrality observed for dual-camera utility.}
    \label{fig:perception_score}
\end{figure*}

Following the trials, participants completed a comprehensive usability questionnaire. The user study aimed to identify the best modality for teleoperating a mobile robot in virtual reality. The three navigation modalities were evaluated in terms of their usability, intuitiveness, and efficiency.
The results strongly validate the system's design choices regarding comfort and feedback utility:
\begin{itemize}
    \item Physical comfort and motion sickness:
Simulator sickness is a critical challenge in VR teleoperation. Our decoupled camera rendering approach proved highly effective, as $95\%$ of participants reported no symptoms of nausea, dizziness, or fatigue after the experiment, and only one participant reported minor eye strain. These results validate the suitability of the proposed system for extended use.
    \item Efficacy of haptic feedback:
Survey responses (see Fig.~\ref{fig:perception_score}) showed overwhelming support for the haptic subsystem. Participants strongly agreed that the vibration feedback helped them avoid collisions and that it was immediately clear why the controllers were vibrating. Participants disagreed with the statement that vibrations were distracting, confirming that the feedback loop provided intuitive cues without increasing cognitive load.
    \item Sensory reliance:
When asked about their navigation strategy in Mode 3, a large number of users agreed with the statement: ``I relied more on the vibration than the video to know when I was too close". This confirms our hypothesis that haptic feedback successfully offloads the visual channel, allowing operators to perceive obstacles in their blind spots without constantly scanning the video feed.
\end{itemize}

\section{Performance Analysis \& Operational Insights}
\label{sec:performance}
The experimental results validate the core hypothesis that multimodal sensory feedback significantly enhances teleoperation performance, particularly in restricted environments. Beyond the raw metrics of time and collisions, several key insights emerge regarding operator behaviour and system applicability.

\subsection{Mitigation of Cognitive Load}

A primary challenge in remote teleoperation is the high cognitive demand placed on the operator, who must mentally reconstruct a 3D environment from 2D video feeds. The qualitative feedback from our study suggests that the proposed haptic system successfully offloads this demand. By providing directional haptic cues, the system effectively externalizes the collision-check process. Operators no longer need to constantly scan the periphery of the video feed for obstacles; instead, they can focus their visual attention on the global path planning (via the map) and forward trajectory, relying on the haptic interface to monitor the robot's immediate vicinity. This sensory substitution is likely the driving factor behind the $23\%$ improvement in navigation speed, as operators felt safer maintaining higher velocities in narrow corridors.

\subsection{Bridging the Skill Gap}

One of the most significant findings of this study is the system's ability to democratize expert-level performance. In the baseline modality (Mode 1), the performance variance was high, with novice users struggling significantly with depth perception and spatial awareness. The introduction of haptic feedback (Mode 3) drastically reduced this variance (as evident in Fig.~\ref{fig:graph}), allowing users with no prior robotics experience to perform comparably to those with moderate experience. This suggests that the multimodal interface acts as a skill equalizer, making it highly suitable for rapid-deployment scenarios, such as search and rescue or disaster response, where operators may need to control complex machinery with minimal training.

\subsection{Implications for Low-Visibility Scenarios}

While our experiments were conducted in a well-lit indoor environment, the reliance on occupancy grids and LiDAR-based haptics provides significant advantages in low-visibility conditions. In real-world scenarios involving smoke, fog, or darkness, standard RGB cameras (Mode 1) exhibit limited functionality. However, the proposed system’s reliance on LiDAR for both mapping (Mode 2) and haptic feedback (Mode 3) implies that navigation capabilities are preserved even if the visual input is compromised. This redundancy makes the architecture particularly robust for subterranean or hazardous environments.

\subsection{Limitations}

Despite the demonstrated success, the current system relies on a 2D occupancy grid, which assumes a structured environment with vertical walls. This approach poses limitations when encountering obstacles that do not appear at the LiDAR’s scan height, such as tables with overhanging tops or uneven terrain. This limitation, observed in the ``stop-and-check" behaviour in Mode 2, highlights the necessity for 3D volumetric mapping. 
Although separating the camera view helped prevent motion sickness, it caused a temporary sense of disorientation. New users required a short time to adapt to the difference between where they were looking and where the robot was moving.

\section{Conclusion \& Future work}

This paper presented a robust, multimodal XR-based teleoperation framework designed to address the inherent limitations of conventional visual-only interfaces. By integrating a stabilized dual-camera stream, a real-time 2D occupancy grid map, and a directional LiDAR-based haptic feedback, the proposed system significantly enhances operator situational awareness while mitigating simulator sickness.

Our experimental validation with 20 participants demonstrated that the addition of haptic feedback serves as an intuitive sixth sense, reducing the cognitive load required for obstacle avoidance. The quantitative results show that the full multimodal system (Mode 3) improved navigation efficiency by $23\%$ and reduced collision frequency by $60\%$ compared to the baseline. Notably, the system proved particularly beneficial for novice operators, effectively narrowing the performance gap between inexperienced users and experts. Furthermore, the decoupling of the camera view from the operator’s head pose successfully eliminated motion sickness symptoms in $95\%$ of trials, validating the ergonomic viability of the system for extended operations.

While the current system establishes a strong foundation for immersive teleoperation, several key enhancements can be made to further expand its capabilities: 
\begin{itemize}
    \item 3D map construction and local costmaps: We are currently extending the mapping module to support 3D volumetric mapping \cite{b17}. This will enable the operator to perceive vertical obstacles, such as overhangs or uneven terrain, that a 2D occupancy grid might miss, and will facilitate the generation of local costmaps to support more intelligent path planning.
    \item Dynamic obstacle filtering: To improve map clarity in populated environments, we are developing dynamic filtering algorithms. This will differentiate between static infrastructure, e.g., walls, and moving agents, e.g., humans and robots, preventing transient objects from leaving permanent ghost obstacles on the navigation map.
    \item Multi-sensor fusion: To address the minimum range limitations of LiDAR (near-field blind spots), we are studying the addition of ultrasonic or infrared proximity sensors. Fusing this data will provide redundant safety checks for objects in immediate proximity to the robot.
\end{itemize}

Beyond these enhancements, the proposed framework could also be integrated with Vision-Language Models. By utilizing the headphones and microphone integrated into the VIVE headset, the addition of VLM-based tools can allow operators to request semantic descriptions of the scene from the robot \cite{b25}, or to interact with the surroundings through natural language. Such an integration can further benefit from robot-agnostic control capabilities, enabling commands to be directed to specific robots through a centralized and robot-agnostic architecture \cite{b23}.

\section*{Acknowledgment}
The authors would like to thank Edoardo Del Bianco for designing and fabricating the custom mounting frames for the HMD base stations and the robot’s camera system. The authors also sincerely thank the 20 participants who took part in the user study for their time, effort, and valuable feedback.
This work was supported by the Liguria Region and Leonardo S.p.A. under Grant LDO/INN/P/0022707/24.

\vspace{12pt}
\color{red}

\end{document}